# A Neural Architecture for Detecting Confusion in Eye-tracking Data

**Shane Sims** and **Cristina Conati**
Department of Computer Science
University of British Columbia
{ssims, conati}@cs.ubc.ca

## Abstract

Encouraged by the success of deep learning in a variety of domains, we investigate a novel application of its methods on the effectiveness of detecting user confusion in eye-tracking data. We introduce an architecture that uses RNN and CNN sub-models in parallel to take advantage of the temporal and visuospatial aspects of our data. Experiments with a dataset of user interactions with the ValueChart visualization tool show that our model outperforms an existing model based on Random Forests resulting in a 22% improvement in combined sensitivity & specificity.

## 1 Introduction

There is increasing interest in creating AI agents that can predict their users' needs, states, and abilities, and then personalize the interaction accordingly, including understanding and reacting to a user's affective state. One such state is *confusion*, which is particularly relevant to user experience while interacting with complex interfaces because when a user is confused, they can experience a decrease in satisfaction and performance [e.g., Nadkarni and Gupta, 2007]. A system that can detect its user's confusion gains an awareness that can be leveraged to provide appropriate interventions to resolve such confusion. Detecting and resolving confusion is becoming especially relevant in supporting users interacting with Information Visualization because data visualizations are now widespread in our daily lives, and confusion has been found to hinder their usage especially when they increase in complexity [e.g., Lee *et al.*, 2016].

Prior work [Lallé *et al*., 2016] showed that confusion during visualization processing can be detected using a Random Forest (RF) classifier and features based on summative statistics of eye-tracking (ET) data (user gaze, pupil size, and head distance from the screen) computed as the interaction unfolds. This classifier achieved 57% and 91% accuracy in predicting confusion and lack thereof, respectively. In this paper, we investigate whether we can improve upon the results of [Lallé *et al*., 2016] by employing a deep learning model to detect confusion from the same ET data set.

The use of deep learning is generally limited in research on modeling user affect, partially due to the difficulty in collecting and labelling large amounts of relevant data. Corpora of data are available for sentiment analysis, i.e. detecting positive vs. negative affect (valence) from text, because it is relatively easy to label for valence, at least as compared to generating labels for finer-grained emotional states. There has been work in using deep learning to detect affect from acted emotions in video [e.g. Ebrahimi *et al.,* 2015], where the affective labels are known a priori. By comparison, collecting datasets for specific unscripted user affective states in interactive tasks is very laborious, and thus such datasets are usually small compared to those in domains where deep learning has been most successful [e.g. Karpathy and Fei-Fei, 2015].

For this reason, approaches to predicting user affect mostly use classical machine learning methods similar to [Lallé *et al*., 2016], with a few exceptions such as [Jiang *et al.,* 2018; Botelho *et al.,* 2017; Hutt *et al.,* 2019]. All of these works seek to predict multiple emotions (including confusion) in students interacting with educational software. They leverage Recurrent Neural Networks (RNNs) to learn from sequences of student interface actions but do so with engineered features based on knowledge of what is important while interacting with each system, thus not fully leveraging the power of RRN to learn representations from low-level data.

Data scarcity is exacerbated with ET data as it currently requires specialized equipment and collection in a lab setting. The dataset used in this paper is no exception, containing data from only 136 users. Yet, we present a deep learning architecture (section 4) that achieves a 49% improvement in detecting confusion compared to [Lallé *et al*., 2016], with no major loss in detecting an absence of confusion (section 5.2).

Thus, one contribution of this work is that, to the best of our knowledge, we are the first to show the suitability of a deep learning approach for the task of classifying affect from ET data. This result can have wider implications for the use of ET data in user modelling as a whole, where such data has been shown to have great potential for modelling affect [e.g. Jaques *et al.,* 2014; Lallé *et al.,* 2018; Bixler & DMello, 2015], as well as cognitive abilities [e.g. Kardan & Conati, 2015], and long-term traits [e.g. Steichen *et al.,* 2014]. By demonstrating the effectiveness of using deep learning based methods with a small eye-tracking dataset, we hope to provide an impetus for further research in this direction.

Our second contribution is the architecture we designed to achieve our results, which combines a Recurrent Neural Network (RNN) and a Convolutional Neural Network (CNN) to learn from sequential and visuospatial information in the ET data. Previous work has combined CNNs and RNNs by using the output of the first as the input to the second (see Related Work below), which is suitable for processing video-based data. Unlike videos, our data has the unique property that a temporal sequence of ET data can be represented in a single frame in a meaningful way, namely with a *scan path* image recording spatial information about the aggregate eye-movements in the sequence. In contrast, such an aggregation of many video frames into one is meaningless. To leverage this property of ET data, our proposed architecture (section 4.3) includes an RNN and a CNN that operate in parallel, with the RNN taking as direct input sequential raw eye-tracking samples (section 4.1) while the CNN processes the corresponding scan path image (section 4.2). The models are trained together in an end-to-end fashion as one unit. A formal evaluation of the model (section 5.3) shows that it achieves better performance than either of its components do alone. We see our results as promising evidence that our proposed approach is worthy of further investigation as more data of this type becomes available and as interest in detecting user states from eye-tracking data continues to increase.

## 2 Related Work

The body of work in predicting affect with deep learning methods is relatively small (compared to tasks like image classification) and occurs mostly in computer vision and NLP, where established methods can be adapted for classifying emotion from images, video, and sound [e.g. Ebrahimi *et al.*, 2015; Xu *et al.*, 2015; Amer *et al.*, 2014]. Exceptions pertain to classifying emotions of students interacting with an intelligent tutoring system [Jiang *et al.*, 2018; Hutt *et al.*, 2019; Botelho *et al.*, 2017]. These works use RNNs to classify emotion from sequences of high-level interaction events (e.g., 'viewing graded questions', 'video play') which may not take full advantage of the RNNs ability to learn a representation from low-level data (e.g. mouse movements).

Eye-tracking (ET) data has shown to contain good predictors of affective and attentional states, such as mind-wandering, boredom and curiosity while interacting with educational software [e.g., Bixler & DMello, 2015, Jaques *et al.*, 2014], and user confusion while interacting with a visualization-based decision support tool [Lallé *et al.*, 2016]. This latter work predicted confusion by combining eye-tracking and interaction data as input to a Random Forest (RF) classifier. The classifier learns from engineered features based on summative statistics (e.g. mean, $\sigma$) of measures related to the user's gaze, pupil size, and head distance to the screen. These measures include, for instance, rate and duration of fixations (gaze maintained at a point), and length and angles of saccades (paths between fixations). We compare our approach based on deep learning directly to this work.

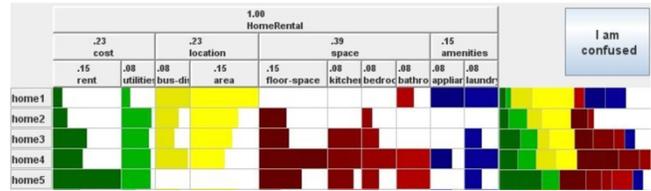

*Figure 1: An example of the main elements of ValueChart*

The only work we identified that uses a deep learning approach to make predictions from ET data aims to diagnose patient developmental disorders [Pusiol *et al.*, 2016]. An RNN is used to learn patterns pertaining to the disorder from how patients look at a trained practitioner who is conducting a diagnostic interview. The model takes as input temporally developed scan paths of the patient gaze superimposed on a video of the interviewer's face. Deep learning has also been used for *gaze estimation* (i.e. predicting the (x,y) coordinates of a person's gaze) from images of the viewer's face [e.g. Zhang *et al.*, 2015]. Note that gaze estimation is what eye-trackers do, and it is distinct from using the estimated gaze data for a predictive task.

There is substantial work that (like our own) combines the particular strengths of RNNs and CNNs. These works [e.g. Donaheau *et al.*, 2014; Srivastava *et al.*, 2015; Yu *et al.*, 2017] relate specifically to processing videos, using a model class known as *Recurrent Convolutional Networks* (RCNs). RCNs typically operate on an input of image sequences (i.e. the frames of a video), where at each step a CNN extracts visual features from the given frame and then feeds this as input to the RNN, which models the temporal dynamics of the sequence. All of these RCN-based approaches make *serial* use the CNN and RNN. Our approach differs in that our architecture uses the two models in parallel, to better leverage the unique nature of our data (section 4.3).

## 3 Dataset

The dataset used in this paper is the same one used in [Lallé *et al.*, 2016]. It was generated via a study designed to collect labelled data for episodes of confusion from users interacting with ValueChart, an interactive visualization-based tool for preferential choice decision making. Figure 1 shows an example of ValueChart configured for selecting rental properties from a set of alternatives (listed in the left-most column), based on a set of relevant attributes (e.g., location or appliances, listed at the top). Although extensively evaluated for usability, the complexity of the decision tasks means that ValueChart can still generate user confusion.

In the study, 136 participants performed tasks with ValueChart, relevant to exploring available options for a home rental decision problem. There were 5 task types, each repeated 8 times, resulting in 5440 tasks (mean duration 13.7s, $\sigma$=11.3). The user's eyes were tracked with a Tobii T120 eye-tracker embedded in the study computer's monitor. In addition to gaze position, this eye-tracker also collects information on user pupil size and head distance from the screen.

To collect ground truth labels for confusion, users self-reported their confusion by clicking on a button labelled *I am*

*confused* (top right in Figure 1). The confusion reports were verified by asking users to confirm them after seeing replays of relevant interaction segments. This process resulted in 112 (2%) tasks with reported confusion and 5328 without. Each item in the dataset is a task segment that ends when a confusion self-report occurs, or at a randomly selected point for tasks when no confusion was reported. The last second of data before a confusion report is removed to exclude signs of the intention to push the *I am confused* button.

## 4 Models and Approach

This section describes the intuition behind using an RNN and CNN on ET data and combining them in a way that is appropriate for our data. Because of the relatively small size of our dataset, in each case, it was important to minimize model complexity. Thus, reducing the number of learnable parameters to avoid overfitting was the driving force behind the various design choices described in this section.

### 4.1 RNN

RNNs are especially suited for sequential data, such as ET data. We chose to investigate RNNs because of the nature of confusion itself. As an affective state, confusion doesn't occur instantly. Rather, it develops over a period of time as the brain uncovers discrepancies between its existing knowledge and what is observed, and continues with subsequent attempts to resolve it, until the person either resolves their confusion or gives up [D'Mello *et al.*, 2014]. It is not well understood how confusion depends on events further back in time (long-term temporal dependence) or develops in a strictly local sequence (local temporal dependence). Note that because of a high sampling rate, in ET data "long" could be many samples ago, yet only a second in the past. RNNs are able to handle both scenarios, which is why it was chosen for this investigation.

Two variations of RNN have become popular for modelling temporal data: Long-Short Term Memory (LSTM) networks and Gated Recurrent Units (GRU). LSTMs are gated RNNs that use self-loops to facilitate the learning of long-term dependencies while also ensuring long-term gradient flow [Hochreiter and Schmidhuber, 1997]. A GRU is essentially a simplified LSTM that reduces the number of gates and thus of learnable parameters [Cho *et al.*, 2014]. Because of this reduction in parameters, we chose to use the GRU as the RNN sub-model in our architecture.

The Tobi T120 eye-tracker collects raw eye-tracking samples at a rate of 120 Hz. This raw data is usually processed with proprietary software into sequences of fixations, identified by clustering raw data to distinguishing small eye-movements from real attention shifts. Leveraging fixations and saccades (gaze paths between fixations) is the standard way to analyze ET data. In fact, the results on detecting confusion by Lallé *et al.* (2016), which are the gold standard for our work, showed that summary statistics around fixations and saccades are strong features for classifying confusion. In contrast, we leverage GRU to learn from the *raw ET samples*, the lowest level of data available from the eye-tracker, to ascertain whether the GRU can pick up any further discriminators useful for classifying confusion. Any patterns that could be lost in going to a higher level of data abstraction is necessarily maintained at this level, where the model has the opportunity to discover these patterns, as well as any interactions among them [Bengio *et al.*, 2013].

Each raw ET sample includes, for each eye: (i) the *x* and *y* gaze coordinates on the study screen; (ii) the size of the pupil; (iii) the distance of that eye from the screen. Therefore, each data item is a 2D array with the number of rows corresponding to the number of samples captured in each of the confused/not confused tasks described in section 3.

A difficulty in using raw ET data collected at a high sampling rate as input to an RNN is the length of the resulting sequences. While there is no fixed length on which RNNs must operate, in practice sequences should be shorter than 400 steps [Neil *et al.*, 2016], whereas the sequences in our dataset average 1644 steps. We use two approaches to reduce sequence length. First, we consider only the last 5s of ET samples in each data item, since Lallé *et al.* (2016) found this interval to perform as well as considering the full length of data. The second approach derives from observing that, because of the high ET sampling rate, values change only a small amount from one sample to the next. Thus, we split the sequence of each data item into four separate data items with the same confusion label. We do so by performing a cyclic split (e.g., as when dealing a deck of cards), which preserves the temporal structure of the time series data. This process reduces our sequence lengths by a factor of four

Cyclically splitting the data has the added advantage of providing the models with multiple opportunities to learn from the same example in a more intelligent way than by duplicating an example exactly (random over-sampling), a naive but common augmentation technique [Gong and Chen, 2016]. The difference between resulting items provides intra-class variance, while the cyclic partition ensures the preservation of the data's sequential pattern.

We perform further augmentation to address the imbalance between confused and not-confused data items in the dataset. Lallé *et al.*, (2016) successfully used *Synthetic Minority Oversampling Technique* (SMOTE) [Chawla *et al.*, 2002] for their RF model, but recall that they were not leveraging the temporal nature of ET data. SMOTE is not commonly used to augment sequences, partially because it measures similarity between samples by Euclidean distance, which is a bad match for temporally misaligned pairs [Gong and Chen, 2016]. However, we ran preliminary experiments showing that SMOTE increased RNN performance with our data, possibly because confusion self-reports provide an anchor that maintains a degree of temporal alignment in our sequences. Thus, we employ SMOTE for augmenting data used by the GRU (see section 5.1).

Based on evidence that for RNNs, depth in the traditional sense (i.e. the number of layers) is not as important as recurrent depth for classification tasks [Zhang *et al.*, 2016], we limit our model to a single layer, thus reducing complexity.

We chose a hidden state of 256 units during hyperparameter tuning using common heuristics [Goodfellow *et al.*, 2016].

## 4.2 CNN

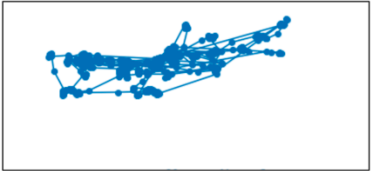

*Figure 3: example scan path generated from raw samples*

Another way in which a sequence of raw ET samples can be represented is as a *scan path image*. Given the coordinates in the raw eye-tracking samples, these images are created to contain the path made by the user's gaze over the sequence[1] (see Figure 3). We leverage a CNN architecture to predict confusion from the scan path images of the ET sequences in our dataset. Note that because of the length of the sequence does not change the size of the corresponding scan path image, we use full sequence lengths as CNN input (as opposed to 5sec segments as done for the RNN), so as to leverage full information of the user's gaze activity over the trial.

Using a CNN to learn from a scan path image is to some extent less intuitive than using an RNN to predict confusion from the raw ET sequences. The scan path is readily represented with an image, but this is not the same as the images that CNNs are typically used for. For instance, natural images contain a hierarchy of parts (e.g. a car's wheels and their subcomponents), as well as properties such as colour and texture, that CNNs have been shown to model in their various layers [Girshick *et al.*, 2014]. No such hierarchies nor properties appear in a scan path image, which is a strictly spatial representation of gaze data consisting of dots and the lines connecting them. In a scan path image, the temporal information on the gaze sequence is lost, but spatial information comes to the forefront. While we cannot identify when a given gaze sample occurred relative to the beginning of the sequence, we can readily see where it is located in relation to the others. And while we cannot identify when the user looked at a specific area, the density of gaze points provides an indication of user attention to that area, while the lines connecting the gaze points indicate the relative length and frequency of the saccades to and from that area.

A CNN can capture these relevant scan path characteristics. Shared weights across a given channel of a convolutional (conv) layer may capture spatial information such as the position of dots and lines, and while deep networks result in spatial invariance (due to pooling), more of this information is maintained in our simple two-layer network. Local information in the image, such as density of dots, is captured in individual neurons related to a portion of the image determined by the model's receptive field (the chain of connections linking a neuron to the patch of an image it has "access" to). We chose some hyperparameters of the CNN with this knowledge of our data and the CNN model class in mind while balancing the competing goal of minimizing learnable model parameters to prevent overfitting our small dataset. As scan paths consist of dots and lines, the deep hierarchies associated with natural images are not required. As such, our CNN consists of two conv layers, which we determined during hyperparameter tuning (section 5.1) by increasing from one until validation set performance decreased. This number makes sense as two layers are enough to extract simple visual features while avoiding the additional parameters that come from unnecessary layers. Having so few layers is also crucial to avoid learning patterns unique to the training data, however, it prevents the model from building a large receptive field (important for capturing local information) via depth. To balance this, we use a slightly larger kernel size (5x5 vs. the more common 3x3) to increase the receptive field's width directly, an acceptable solution for our model despite the cost of more shared weights. Next, colour has no meaning in a scan path image, so we use a single grayscale input channel. As our images do not contain fine or nuanced textures (like the hair of an animal for instance), high resolution is not important, allowing us to downsize our images by a factor of 6, thus reducing the dimensions and parameters of each conv layer.

---
[1] Such images are commonly available via the eye-tracker's software, based on fixations.

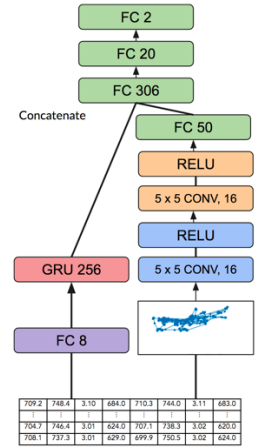

*Figure 4: VTNet Architecture. We use the TYPE-size convention for labelling layers. Here FC represents a fully connected layer.*

To finalize the model's architecture, we chose the number of filters in each conv layer (6 and 16 for the first and second layer, respectively) during hyper-parameter learning, using common heuristics [Goodfellow *et al.*, 2016].

Note that, because we use the full ET sequences in the scan path, having substantially longer sequences of variable length, preclude the use of SMOTE as an augmentation technique [Gong and Chen, 2016] (more on this in section 5.1).

## 4.3 VTNet

Having developed the intuition behind using each of RNNs and CNNs on eye-tracking data, here we describe an architecture to leverage the strength of both together. Previous architectures that combine CNNs and RNNs (see section 2), do so serially at each point of the temporal data (e.g. video frames) and are thus not suitable for our learning task. Processing a scan path image as it develops over time through a CNN to extract features for RNN input gives no more information than that already available in the raw sequence.

Instead, in our approach, each of the CNN and RNN takes a different representation of the same data sequence and processes it independently. The output is then combined and given as input to fully connected (FC) layers for classification. This approach is well suited to our particular data. Unlike videos, ET data has the unique property that a given temporal sequence of data can be represented in a single frame in a meaningful way. That is, a given image of a user's entire scan path contains information about the aggregate spatial eye-movement (while an aggregation of many video frames into one is meaningless). In addition to this, an RNN looks at how a scan path develops over time.

Our hypothesis was that having a model that can process this multimodal data representation will enhance its predictive abilities. This model (visuospatial-temporal model or *VTNet* from now on) is shown in Figure 4. It combines an RNN and a CNN to model, respectively, sequences of raw ET samples and scan paths, as described in the previous sections. These two sub-models run in parallel with their distinct inputs. To learn a shared representation of the temporal and visual-spatial input data, the output of the CNN and RNN are concatenated into a single vector, fully connected to a simple neural network with one hidden layer, which classifies the input as either *confused* or *not confused*. The entire model is then learned end-to-end as a single unit.

### 4.4. Implementation

All of our neural network-based models are implemented and trained using PyTorch (code available on GitHub at *anonymous for review*). We use negative log-likelihood as our loss function, with the Adam optimizer [Kingma and Ba, 2014]. We limit training to 100 epochs, employing linear learning rate decay and early stopping to end training when validation performance stops improving. We train our models using a single Nvidia GTX 1080 GPU.

## 5 Evaluation

Our evaluation is designed to first determine how the GRU model (the RNN component of VTNet) described in section 4.1 performs compared to the implementation of the RF approach in [Latte et al 2016] (section 5.2). We begin with this comparison because the RNN is the most intuitive neural network model to use with raw ET data. We then determine whether combining the RNN with a CNN in the proposed VTNet architectures is more effective than its constituent parts are alone (section 5.3).

### 5.1 Experimental set-up

Model performance is evaluated with *sensitivity* and *specificity*, which are the proportion of *confused* and *not confused* tasks correctly identified as such, respectively. Because of the dataset's class imbalance, both metrics together are more meaningful than accuracy alone. For instance, a 98% accuracy could be achieved by simply classifying everything as not confused, but not capturing any instance of confusion. We also report the mean of sensitivity and specificity scores as a unified measure of performance (*combined accuracy*).

All models are evaluated using 10 runs of 10-fold cross-validation (giving 100 iterations of CV in total) to reduce fluctuations in the results due to the random selection of folds. All results reported in the next section are the average of the 10 runs of 10-fold CV. Cross-validation is done across users. Namely, no user contributes data points to both the training and test sets of a given fold, thus measuring model performance on unseen users. Cross-validation is also stratified, i.e., the distribution of confusion data points in each fold is kept similar to that of the dataset as a whole.

For the RF model, nested CV (i.e., further cross-validation on each training set) was used for feature selection, hyperparameter tuning, and to choose the decision threshold that maximizes sensitivity and specificity[2]. For the deep learning models, using nested CV would be computationally too onerous. Instead, for each of the 100 iterations of CV we randomly select 20% of the data to serve as a validation set for hyperparameter tuning and decision threshold setting. Note that contrary to the nested-CV, the validation set is holdout data that is not re-added to the training set for training prior to evaluation on the test set. This effectively results in the DL models being trained on 20% less data than the RF model. For the RRN (when used on its own) and RF models, classes in the training sets are balanced by first using SMOTE to increase the size of the minority class (confused) by 200% and then randomly down-sampling the majority class (as was done in [Lallé *et al.,* 2016], resulting in approx. 1350 confused data items. SMOTE cannot be used with the CNN (section 4.2), nor with the VTNet that includes it. Thus, for these models, we just down-sample the majority class to achieve class balance, resulting in approx. 450 confused items. Validation and/or test sets are left unbalanced in all models.

| Model | Sens. | Spec. | Combined |
|---|---|---|---|
| RF | 0.53 | **0.80** | 0.67 |
| GRU | **0.75** | **0.80** | **0.78** |

*Table 1: Test set performance of GRU and RF.*

### 5.2 Results of Comparing GRU and RF

The result of our first comparison is summarized in Table 1, which shows that GRU outperforms the RF classifier in both sensitivity and combined accuracy, with no change in specificity. The GRU achieves a combined accuracy of 0.78 ($\sigma = 0.01$), compared to the 0.67 ($\sigma = 0.05$) achieved by the RF. We test this result with an independent samples t-test, which shows that the difference is statistically significant[3] ($t_{18} = $ 6.28, $p < .001$). The difference in sensitivity also proves significant ($t_{18} = 6.22, p < .001$), with a substantial 41.5% improvement over the sensitivity of the RF model. These results allow us to conclude that the GRU outperforms the RF in classifying confusion with this dataset, where the impact of the GRU is specifically improving sensitivity, namely detecting confusion when it occurs, with no loss in the accuracy of predicting when a user is not confused.

---

[2] This is done by choosing the threshold closest to the (0,1) point on the Receiver Operating Characteristic (ROC) curve.

[3] Significance is defined at p <.05 throughout the paper

Lallé *et al.,* (2016) also experimented with features based on interaction events (e.g., the user clicking an interface element). Combining these with eye-tracking features gave them their best results (0.61/0.926 sensitivity/specificity), for a combined accuracy of 0.768. With this additional data modality, the RF still doesn't perform better than the GRU trained only on ET data. This result is especially encouraging when we consider that the GRU is trained on 20% less data (the portion held out as the validation set).

### 5.3 Results of Comparing VTNet, RNN and CNN

| Model | Sens. | Spec. | Combined |
|---|---|---|---|
| GRU | 0.75 | 0.80 | 0.78 |
| CNN | 0.73 | 0.80 | 0.77 |
| VTNet | 0.79 | 0.84 | **0.82** |

*Table 2: Test set performance of neural models.*

After establishing the superiority of the GRU vs. the RF model in classifying confusion, we evaluate if the performance of the GRU can be improved by combining it with a CNN into the VTNet architecture described in section 4.3. The result of this comparison is summarized in Table 2. The VTNet has been trained with the same hyper-parameter configuration as its corresponding sub-models. We see that for all three measures (sensitivity, specificity, and combined accuracy) the VTNet outperforms both the GRU and the CNN. A one-way ANOVA with classifier type (VTNet, GRU, and CNN) as the factor shows a significant effect on all three measures. Post hoc testing via Tukey HSD (which adjusts for multiple comparisons) shows that for all three measures, the difference is statistically significant between VTNet and both GRU and CNN, with no significant difference between the latter two. With this we conclude that VTNet surpasses the performance of both of its constituent parts and is thus an effective model for classifying confusion from our ET data.

VTNet achieves a 79% accuracy in sensitivity, which represents a 49% increase over the original RF model. It is also the only one of the three deep learning models to increase sensitivity (reaching 84%), suggesting that combining temporal and visuospatial information on gaze manages to capture patterns pertaining to the absence of confusion that go otherwise undetected. That neither the VTNet nor the CNN has SMOTE augmented data, yet still outperform the GRU with augmented data, indicates that there is a strong signal for confusion in the scan path images that is not obvious. Understanding this signal worthy of further investigation.

The performance of the VTNet model is also higher than other published approaches to predicting confusion using RNNs in a different context, namely leveraging the interaction data of users while they study with ITSs [Jiang *et al.,* 2018; Botelho *et al.,* 2017]. Neither work reports sensitivity nor specificity, but both report AUC for the models ROC. The prior reports an AUC of 0.57, the latter an AUC of 0.72 By comparison, we achieve an AUC of 0.84 with VTNet, although it is possible that our specific combination of task (processing an InfoVis) and input data (eye-tracking) provides better conditions for detecting confusion.

## 6 Conclusions and Future Work

In this paper, we presented a novel approach that leverages deep learning for detecting user confusion from raw sequences of eye-tracking (ET) data. The approach combines the CNN's strength in spatial reasoning with the RNN's strength in temporal reasoning, and the resulting model (VTNet) outperforms either of these methods considered on their own, when tested on a dataset of users interacting with a complex visualization for decision support. VTNet also largely outperforms a previous model based on Random Forests, on the same dataset [Lallé et at 2016]. Our approach extends previous work on combining CNNs and RNNs for processing video images. This is done in a manner that suits the specific sequential and visuospatial nature of the ET data, where a temporal sequence of raw samples in a given timeframe can also be represented as a single visual scan path for that timeframe.

Our results represent a 22% increase in combined sensitivity and specificity, with the bulk of the increase (49%) being in sensitivity (79% accuracy in detecting confusion when it occurs), which is remarkable considering that our dataset contained only 2% datapoints for confusion. While deep learning brought about 16-23% improvements when initially applied to speech recognition and a 41% reduction in error rate when applied to object recognition [Bengio *et el.*, 2013], both of these domains have much more available data.

As future work, we will explore methods for increasing the performance of the VTNet, such as increasing receptive field size via dilated convolutions. We also intend to combine interaction data with the ET data, as done in [Lallé *et al.,* 2016]. Finally, the generality of VTNet should be tested on other ET datasets that have been used to predict user states other than confusion (e.g. learning, affective valence, visual abilities).